\documentclass[10pt,twocolumn,letterpaper]{article}

\usepackage{iccv}
\usepackage{times}
\usepackage{epsfig}
\usepackage{graphicx}
\usepackage{amsmath}
\usepackage{amssymb}
\usepackage{my_defs}

\iccvfinalcopy 

\def\iccvPaperID{****} 

\ificcvfinal\fi

\newcommand\blfootnote[1]{%
  \begingroup
  \renewcommand\thefootnote{}\footnote{#1}%
  \addtocounter{footnote}{-1}%
  \endgroup
}

\begin{document}

\title{Semantic Scene Completion Combining Colour and Depth:\\ preliminary experiments}

\author{Andr\'e B. S. Guedes ~ ~ ~ ~ Te\'ofilo E. de Campos\\ 
Universidade de Bras\'{\i}lia\\
Bras\'{\i}lia-DF, 70910-900, Brazil\\
{\tt\small t.decampos@st-annes.oxon.org} \\ 
{\tt\small \url{http://www.cic.unb.br/~teodecampos/}}\\
\and
Adrian Hilton\\
CVSSP, University of Surrey\\
Guildford, GU2 7XH, UK\\
{\tt\small a.hilton@surrey.ac.uk}
}

\maketitle

\begin{abstract}
  Semantic scene completion is the task of producing a complete 3D voxel 
representation of volumetric occupancy with semantic labels for a scene 
from a single-view observation. 
We built upon the recent work of Song et al.~\cite{song_etal_SSCnet_cvpr2017},
who proposed SSCnet, 
a method that performs scene completion and semantic labelling in a 
single end-to-end 3D convolutional network. 
SSCnet uses only depth maps as input, even though depth maps are usually 
obtained from devices that also capture colour information, such as 
RGBD sensors and stereo cameras. 
In this work, we investigate the potential of the RGB colour channels 
to improve SSCnet. 

\end{abstract}
\blfootnote{The work described in this {\em extended abstract} and in the attached poster was presented at the ICCV 2017 Workshop on 3D Reconstruction meets Semantics (3DRMS).}


\section{Introduction}

The task of reasoning about scenes in 3D is one of the
seminal goals of Computer Vision~\cite{marr_Vision_book1982}.
If the 3D geometry of a scene is known, robots are able
to plan trajectories, avoid collisions or clean surfaces.
If the semantic labels of each surface or voxel is also
known a robot can also figure interact with the environment
and perform more complex tasks, such as moving objects
from one location to another; opening/closing
doors, drawers, windows; operating kitchen appliances etc.
Three-dimensional maps with labelled voxels have several
other applications, including surveillance, assistive computing,
augmented reality and so on.
One issue is that capturing the full geometry of a scene can
be time consuming (if it is done using a
scanning technique~\cite{newcombe_etal_KinectFusion_ismar2011})
or expensive (if it is done using a rig of calibrated sensors).

It is well known that vision is a combination of so called
bottom-up and top-down processes~\cite{marr_Vision_book1982}.
Bottom-up information can be obtained by matching local features
for stereopsis and top-down is the use of prior knowledge from
related scenes and objects.
If both types of cues are combined, it is possible to estimate
a complete scene geometry by using a single visual and depth
map of a scene. This is well illustrated in
Figure~2 of~\cite{song_etal_SSCnet_cvpr2017}.
A visual sensor captures a single view of a scene
which provides measurements (e.g. RGB and Depth) of the visible objects
but it is not possible to measure the geometry of occluded regions.
However, if the class of the objects is identified, it is possible to
infer the complete scene geometry, enabling a full 3D representation to
be proposed.

Solid computational demonstrations of this
have started to be published recently.
Notably, Song et al.~\cite{song_etal_SSCnet_cvpr2017} introduced the
problem of Semantic Scene Completion (SSC), i.e., given an depth map,
the goal is to generate a 3D image where each voxel is associated
to one out of $N+1$ labels, where there are $N$ known object labels
plus an `empty space' label.

In \cite{song_etal_SSCnet_cvpr2017}, this problem is approached
using a Deep 3D Convolutional Neural Network coined SSCNet.
That paper demonstrates impressive results on completing and
labelling a full 3D scene generated from a single depth map.
Using a combination of bottom-up dues (from the depth sensor)
and top-down cues (learnt from the training set), their method
is able to infer the geometry and labels of the whole scene,
including heavily occluded regions, such as the regions
under tables and behind sofas, as illustrated in
Figure~1 of \cite{song_etal_SSCnet_cvpr2017}.

However, one of the main limitations of SSCNet is that
it was not designed to use any colour information, only
depth maps are used. This clearly impairs the method
as indoor scenes generally include various sources
of error in depth and geometry estimation.
Highly reflective scenes with glass, mirrors or shiny
surfaces usually induce false depth.
If depth is captured using stereo cameras, 
texture-less and non-Lambertian surfaces often result in
errors in feature detection and matching.
Colour information also disambiguates between
different objects that have similar shape or
that are co-planar, like posters on the wall.
Furthermore, it is clear that colour offers
crucial information for semantic labelling that strongly
complement depth information, as seen in
papers that focus on semantic segmentation from
RGBD images, e.g.~\cite{silberman_fergus_SceneSegmentation_iccvw2011,
  silberman_etal_eccv2012, kahler_reid_iccv2013,gupta_etal_ijcv2014,
  wang_etal_tip2015,chen_etal_SceneModelingRGBD_survey_cvm2015,
  eigen_fergus_iccv2015}.

In this paper, we propose to use colour in addition to depth
for Semantic Scene Completion.
For that, we propose modifications of the SSCnet
architecture in order to fuse RGB and depth.
A new input layer was proposed to encode colour
in the visible frustum and we combined a feature
extraction training technique for multiple view learning.


\section{Colour SSCNet}
\label{sec:colour_sscnet}

Depth maps are acquired using an RGB-D sensor and using
the sensor's intrinsic calibration parameters,
a 3D point cloud is generated. The observed geometry
is then encoded using flipped Truncated Signed Distance
Function (fTSDF), proposed
in  \cite{song_etal_SSCnet_cvpr2017}. This method
associates a value to each point in the 3D space to a
function of its distance to the nearest surface point.
The sign of this value indicates if it is visible or occluded.
Apart from the occlusion coding, this method is viewpoint
independent.

The fTSDF encoding of voxels describe the geometry of the
space, but it does not carry any information about the
colour or grey level of the visible objects.
We propose to encode the RGB values of the visible surfaces
in another voxel representation of the scene.
The three channels are normalised to range from 0 to 1.
Empty spaces and occluded regions are coded with the -1
value for the three colour channels.

We apply these two encoding techniques to
RGB and Depth signals and run them through a 3D CNN
that learns to map from RGBD to a labelled 3D volume.
Labelled volumes were obtained as described
in~\cite{song_etal_SSCnet_cvpr2017}, i.e.,
the binvox voxelisation technique~\cite{nooruddin_turk_BinVox_tvcg2003}
was applied to 3D models, which accounts for both
surface and interior voxels using a space carving approach.

We built upon the 3D CNN architecture of
SSCNet~\cite{song_etal_SSCnet_cvpr2017}.
To combine RGB and Depth, we propose the two fusion
schemes described below.

\begin{figure}
  \centerline{
    \includegraphics[width=\columnwidth]{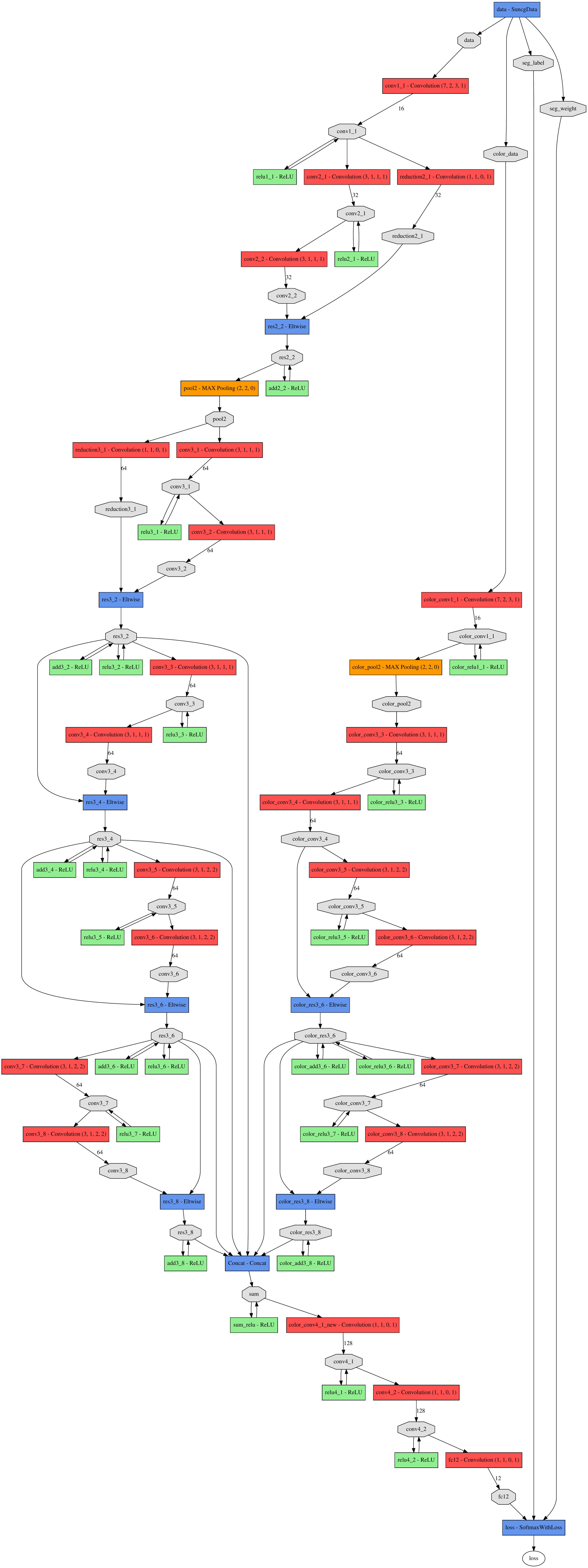}
  }
  \caption{\label{fig:network_architecture}Network architecture for
    mid-level fusion.}
\end{figure}

{\bf Early fusion.}
The first layer of SSCNet was adapted so that it takes as input
a concatenation of fTSDF and the three colour channels
encoded as described above.
The remaining of the network is the same.

{\bf Mid-level fusion.}
This architecture is depicted in 
Figure~\ref{fig:network_architecture}, drawn using
Caffe~\cite{jia_etal_Caffe_arXiv2014} (better viewed
on a screen).
The numbers in brackets are: kernel size,
stride, pad and dilation factor, respectively.
The branch on the left is essentially a copy of
SSCNet. A colour 3D CNN was built following a similar architecture
to SSCNet up to the concatenation layer.
This
layer originally aggregated the output of five scales gathered
from  previous layers and it is followed by three 3D convolutional
layers.
Although this network does not have any fully connected (FC)
layer, the last three layers perform the same class as FC layers
in classification CNNs, as they are closer to the output, which
produces labelled data. A typical
mid-level fusion using CNNs is done at the input of the
first FC layer. Therefore, we believe the most appropriated
layer to fuse SSCNet-like sub-networks is at the concatenation layer.

However, the original SSCNet is very memory-intensive
and it requires about 7GB of GPU memory to process
a single depth image\footnote{The original implementation from
  the authors, obtained from \url{https://github.com/shurans/sscnet},
  actually requires almost 12GB. We removed some redundancy from
  their code, freeing about 5GB of memory.}. 
If we were to duplicate all layers of the network up to the
concatenation layer, much more than 12GB would be required,
whereas most GPU models available nowadays have up to 12GB
of RAM.
To add to the challenge, our RGB coding uses three
channels per image, rather than one as in the fTSDF model,
though this only affects the first convolutional layer.
Therefore, some of the convolutional layers were removed
from the colour branch of SSCNet, but we have preserved all the
dilated convolution layers, as this is a significant
feature of SSCNet which widely expands the receptive
field of the network~\cite{yu_koltun_DilatedConv_iclr2016}.
\fixme{Insert figure 1 of \cite{yu_koltun_DilatedConv_iclr2016}?}

In addition, we also evaluated a {\bf colour-only SSCNet},
which follows the same architecture as the colour branch of the
mid-level RGB-D fusion network, but it is followed by the
top three convolutional layers, without aggregating
activations from the fTSDF branch.

\section{Experiments and Training Strategies}

Our evaluations focused on the NYU depth v2
dataset~\cite{silberman_etal_eccv2012}, using the standard
split of 795 training samples and 654 test samples\footnote{We used the
  train+validation split for training and the test split for testing,
  following the sample indices available from \url{https://github.com/shelhamer/fcn.berkeleyvision.org/tree/master/data/nyud}.}. However, instead of the standard semantic segmentation
labels, we used the labels devised for scene completion, where objects
are grouped into 7 categories plus window, wall, floor, ceiling
and another category that identifies free space.
This set of labels originated from~\cite{handa_etal_SceneNet_cvpr2016}.
As explained in~\cite{song_etal_SSCnet_cvpr2017}, ground truth
volumes were obtained from 3D mesh annotations of \cite{guo_etal_Complete3D_arxiv2015}.
Our implementation was developed using the Caffe framework~\cite{jia_etal_Caffe_arXiv2014}.

We evaluated the two architectures proposed in Section~\ref{sec:colour_sscnet}:
early and mid-level fusion and compared it against the original SSCNet
and colour-only. For all methods that we proposed, training
was done following these strategies: 
\begin{itemize}
\item {\bf Random initialisation}: all parameters were randomly
  initialised and the whole network was trained
  from scratch.
\item {\bf Feature learning}: we kept the original SSCNet parameters trained by
  Song et al.~\cite{song_etal_SSCnet_cvpr2017} for all the original layers
  and optimised only the colour layers, i.e., the original SSCNet
  parameters were frozen.
\item {\bf Fine tuning}: this is similar to the strategy above, except that
  instead of freezing the original layers, we also enabled their parameters
  to be optimised, but with the  learning rate ratio of 0.2 times
  the ratio of the new layers.
\item {\bf Surgery}: was applied only for the early fusion approach. It is similar
  to fine tuning, except that the weights of the input layer which
  related to depth were set to the original parameters of the first layer of
  SSCNet and the other weights (linked to the colour channels) of the same
  convolutional kernel were initialised randomly.
\end{itemize}

Voxel labelling is done by applying soft-max to the scores of the
last convolutional layer of the networks and optimisation is done
using cross-entropy as a loss function, averaged out over all classes.
\fixme{except empty space?}



\fixme{Implementation details:
  \cite{ghemawat_dean_LevelDB_2011,
    chu_lmdb_concept_2011,
    dempster_etal_EM_1977,
    karpathy_CNNweb_2016}
}

The results were evaluated using the Intersection over the Union
(IoU) between predicted class labels and ground truth, averaging out
over all voxels in the test set and all classes.
We followed \cite{song_etal_SSCnet_cvpr2017} and evaluated our
results both in terms of completion (i.e., the ability to
detect if an occluded voxel is occupied or free space) and
in terms of semantic labelling of voxels of all classes.

\section{Results and Discussion}

Our results so far show that none of the proposed architectures and
training strategies actually lead to results that are better than
the original SSCNet based only on depth observations, i.e.,
through the training iterations, our results peaked at
scene completion IoU of 56.6 and average semantic
scene completion of 30.5, which are both results
obtained by the original SSCNet on the test set of the NYU depth v2 dataset.
In other words, our experiments in the NYU depth v2 dataset
(with the 12 category labels~\cite{handa_etal_SceneNet_cvpr2016})
show that the proposed method for coding colour information
is not as discriminative as fTSDF, neither it complements
depth information.

However, the performance of our colour-only network, initialised with
randrom weights, followed a monotonic increase as the number
of training iterations increased, though it did not converge
with the same number of iterations as the architectures
that use depth.
Therefore, there is certainly relevant information in RGB, but
it should probably be combined with fTSDF in a different way,
perhaps using late fusion.
Even if early or mid-level fusion are not the ideal strategies
in this problem, further investigation is also needed to
understand why RGB has not complemented Depth at all. It might
be an artefact of the dataset and set annotated classes,
as it is possible that geometry alone is already very
discriminative. A suggestion is to verify this using
more complex scenes with more occlusions or with finer
object class labels.

Our results have also shown that unconstrained Fine Tuning
leads to a higher decrease in the loss function than the constrained
optimisation methods (Feature Learning and Surgery). However, after
1000 iterations, the test set performance (measured by IoU) starts
to decrease due to over-fitting.
Although the loss is lower for Fine Tuning, we did not observe a
significant difference between the
methods in terms of test set performance.














\section{Conclusion}
In this paper we reported ongoing work that 
considers the problem of Semantic Scene Completion in 3D
from a single RGBD image.
Starting from the 3D CNN architecture of
SSCNet~\cite{song_etal_SSCnet_cvpr2017}, which used only
depth maps as input, we proposed to combine RGB and Depth
information using early and mid-level fusion schemes.

Our preliminary results were not better than the original
depth-only method.
Therefore, further investigation is
needed in order to verify if the dataset (NYU depth v2 with
12 labels obtained from \cite{handa_etal_SceneNet_cvpr2016})
prilidedges structural information such that depth is already
very discriminative. A finer set of labels or a more complex
dataset should be evaluated.
Other directions of future work are to evaluate late fusion
scheme and investigate other ways to encode RGB information.

\section{Acknowledgements}

We are grateful for the valuable comments and suggestions
provided by anonymous reviewers of the first version of
this manuscript.
TEdC's attendance to this workshop is sponsored by 
Funda\c{c}\~ao de Apoio a Pesquisa do Distrito Federal (FAP-DF),
edital 01/2017, protocolo n$^o$18708.76.44500.14072017

{\small
  
}
\begin{figure*}[ht!]
  \vspace{-2cm}
  \centerline{
    \includegraphics[width=1.1\textwidth]{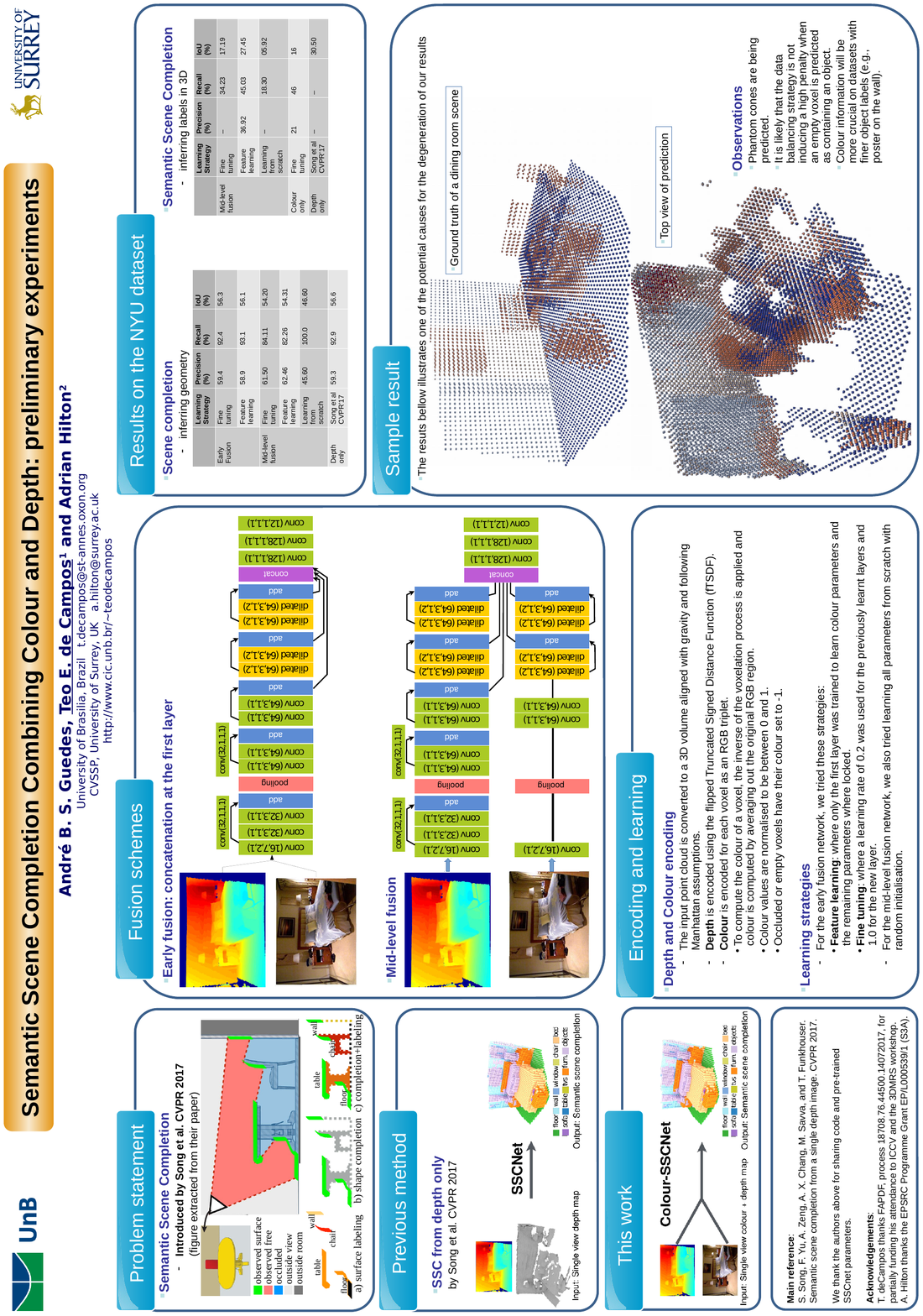}
  }
\end{figure*}

\end{document}